\documentclass[11pt, a4paper, singlecolumn]{article}
\usepackage[utf8]{inputenc}
\usepackage[english]{babel}
\usepackage[top=2cm, bottom=2cm, left=1.9cm, right=1.9cm]{geometry}

\usepackage{titlesec}
\titlespacing{\section}{0pt}{\parskip}{\parskip}
\titlespacing{\subsection}{0pt}{\parskip}{\parskip}
\titlespacing{\subsubsection}{0pt}{\parskip}{\parskip}

\usepackage{indentfirst}
\usepackage[export]{adjustbox}
\usepackage{graphicx}
\usepackage{amssymb}
\usepackage{amsmath}

\usepackage{url}            
\usepackage{booktabs}       
\usepackage{amsfonts}       
\usepackage{nicefrac}       
\usepackage{microtype}      

\usepackage{subcaption} 
\usepackage{adjustbox} 
\usepackage{float} 
\restylefloat{table} 
\newcommand{\tabhead}{\textbf}
\usepackage{multirow}

\usepackage[dvipsnames]{xcolor}

\usepackage[figurename=Fig.]{caption}
\usepackage{caption}
\usepackage[colorlinks, citecolor=ForestGreen]{hyperref}

\usepackage{cite}

\usepackage{abstract}
\title{\textbf{Evaluation Of Hidden Markov Models Using Deep CNN Features In Isolated Sign Recognition\footnote{Preprint of the accepted manuscript at Multimedia Tools  and Applications Journal.}}}
\date{\vspace*{2pt}}
\author{
	\normalsize \textbf{Anil Osman Tur}\\
	\normalsize Ankara University\\
	\normalsize Computer Engineering Department\\
	\normalsize aotur@ankara.edu.tr
	\and
	\normalsize \textbf{Hacer Yalim Keles\footnote{Corresponding author.}}\\
	\normalsize Ankara University\\
	\normalsize Computer Engineering Department\\
	\normalsize hkeles@ankara.edu.tr
}

\begin{document}
\maketitle

\begin{abstract}{
	\vspace*{-1.5em}
	\it Isolated sign recognition from video streams is a challenging problem due to the multi-modal nature of the signs, where both local and global hand features and face gestures needs to be attended simultaneously. This problem has recently been studied widely using deep Convolutional Neural Network (CNN) based features and Long Short-Term Memory (LSTM) based deep sequence models. However, the current literature is lack of providing empirical analysis using Hidden Markov Models (HMMs) with deep features. In this study, we provide a framework that is composed of three modules to solve isolated sign recognition problem using different sequence models. The dimensions of deep features are usually too large to work with HMM models. To solve this problem, we propose two alternative CNN based architectures as the second module in our framework, to reduce deep feature dimensions effectively. After extensive experiments, we show that using pretrained Resnet50 features and one of our CNN based dimension reduction models, HMMs can classify isolated signs with 90.15\% accuracy in Montalbano dataset using RGB and Skeletal data. This performance is comparable with the current LSTM based models. HMMs have fewer parameters and can be trained and run on commodity computers fast, without requiring GPUs. Therefore, our analysis with deep features show that HMMs could also be utilized as well as deep sequence models in challenging isolated sign recognition problem.
	
}\end{abstract}


\section{Introduction}
\label{s_intro}
Recognition of signs/gestures from video streams using computational models aims to find effective solutions to the communication problems between the deaf and the hearing communities. Using machine learning algorithms to generate an effective recognition model for gestures can also help to improve human-machine interaction. A gesture, or sign, is usually a composition of multi-modal sources, such as hand, face and body features. Therefore, the challenge is on following and representing the features of all the source regions, simultaneously. Additional difficulties arise when considering variations imposed by different signers and the environment, such as body and pose variations, and differences in background, illumination, etc. The models need to be invariant to these \cite{escalera_challenges_2017}.

Sign recognition systems are primarily composed of two basic components: (1) representation of the video frame data in a more efficient feature space, (2) classification of the feature sequences of a video stream \cite{pisharady_recent_2015}. Some early works in sign language recognition utilized data gloves \cite{grobel_isolated_1997} or colored gloves \cite{cooper_sign_2012} to track hand movements and deal with occlusion problems more efficiently. However, using external equipment makes these solutions impractical in daily life. Using computer vision algorithms on videos provides more practical solution.

In this research, we propose a generic framework that is composed of three modules to solve the isolated sign recognition problem using deep features effectively with different sequence models, i.e. HMMs and LSTMs. We use the challenging Montalbano dataset \cite{motalbano}, which contains depth and skeleton data in addition to color image data, for our empirical analysis. Without making any explicit hand or face segmentation, our framework takes raw sign videos as inputs and classifies them. Our aim is to provide an effective solution to this problem utilizing both HMMs and LSTMs with deep features. HMMs usually perform as good as deep sequence models, i.e. LSTMs, if the features are robust; they are advantageous, if this is the case, since they have fewer parameters than the deep models and require less data for training.  However, high dimensional feature space create convergence problems in training HMM models; we also experienced this challenge with the deep features in Montalbano dataset. To solve this problem, we included a module in our framework between the feature extraction and sequence classifier modules, to reduce the feature dimensions effectively, i.e. learned from training data, with an additional CNN based model. In this work, we explore different alternatives in each module; in the first module, two pre-trained CNN modules, i.e. Resnet50 and VGG16 models, are evaluated, in the second module Principle Component Analysis (PCA) method and two our custom designed CNN architectures are evaluated, in the third module Gaussian-HMMs, Gaussian Mixture Model-HMMs and LSTM sequence models are evaluated. We performed extensive experiments and provide empirical analysis to solve this problem effectively using HMMs.

The contributions of this paper can be summarized as follows:

(1) We propose a three stage framework that is designed to work with HMMs using deep CNN based features in isolated sign recognition problem from video streams.

(2) We proposed different CNN architectures for dimension reduction, each of which could learn a non-linear mapping from the high dimensional feature space to low dimensional space. Empirical evaluations show that these non-linear projections are more effective than linear projections in this domain. 

(3) We provide extensive experimental analysis with different alternatives in each module of the proposed framework using RGB, depth and skeletal data modalities.

(4) We show that utilizing the same deep features, HMMs can classify sign videos with comparable performances with LSTM models.

The rest of the paper is organized as follows. We provide a summary of the sign recognition problem in Section \ref{s_related}, focusing on Montalbano dataset. We briefly introduce Montalbano dataset and our preprocessing in Section \ref{s_dataset}, followed by our proposed method in Section \ref{s_method}. Finally, we provide the results of our experiments in Section \ref{s_results} and conclude the paper with Section \ref{s_conc}.

\section{Related Works}
\label{s_related}
Sign Language recognition literature is vast \cite{cheok_review_2019}, hence, we will focus more on the literature that is relevant to our approach. 

In sign language recognition, two sequence learning approaches are utilized extensively, HMMs and Recurrent Neural Networks (RNNs). HMMs are suited for temporal data with varying length. In \cite{grobel_isolated_1997}, for example, each sign is modeled with one HMM. For both training and recognition, feature vectors are extracted from each video frame and then fed as inputs to to the HMM. Colored cotton gloves are used to get both trajectory and hand shape features from video frames. In \cite{cooper_sign_2012},  a Kinect camera is used to capture RGB and skeletal modalities to extract appearance-based hand features and track the positions of hands in 2D and 3D space. HMM is used with sequential pattern boosting (SP-boosting). The proposed methods are tested on two datasets: (1) dataset with 20 German Sign Language (GSL) signs and (2) with Kinect 40 GSL signs. They reported 92\% accuracy on dataset (1) with subject dependent testing; 76\% with subject independent testing. On dataset (2) they got 59.8\% with subject dependent testing and 49.4\% with subject independent testing. In \cite{akram_visual_2012}, HMMs are used for isolated sign language recognition with Kinect. Hands are masked with the use of Kinect's skeletal data stream for the hands and the mask that is generated by Kinect. Histograms of oriented gradients (HOG) \cite{hog_method} is used for feature extraction from RGBD data with focus on the hands. The model is tested on the created dataset, which consists of 96 signs. 23 samples are performed by 4 different signers for each sign. They reported 94\% accuracy in the signer dependent case and up to 47\% in the signer independent case. In \cite{jie_huang_sign_2015}, a 3D CNN model is used to solve sign recognition. The performance of CNN model is compared with a  Gaussian Mixture Model (GMM) with HMM, i.e. GMM-HMM, which is chosen as a baseline for comparisons. The used dataset contains a color video stream, depth video stream and the body movements of the signers, simultaneously. In total, they have five types of input data: RGB data, depth and skeletal data. The dataset contains 25 widely used words that are performed by 9 signers with 3 repetitions for each word. Features for the GMM-HMM consist of  HOG features from hand-shape images and body joint locations. They obtained 94.2\% accuracy with the 3D CNN model, and 90.8\% accuracy with the baseline model (GMM-HMM). Reducing the number of modalities, \cite{jan_discriminative_2018} uses minimum classification error training to produce a discriminative HMM classifier using only skeletal features of the Montalbano dataset. The accuracy of this method is 87.3\%, which is comparable to other results reported on the Montalbano dataset using discriminative non-temporal methods. Their research shows that discriminative HMMs can be used successfully as a solution to the problem of isolated gesture recognition. A different approach is presented in \cite{koller_deep_2018}, which introduces an end-to-end embedding of a CNN into an HMM, while interpreting the outputs of the CNN in a Bayesian framework. The hybrid CNN-HMM combines the strong discriminative abilities of CNNs with the sequence modelling capabilities of HMMs.

Similar to the researches mentioned above, in this research we also use HMMs and GMM-HMMs as sequence models for isolated sign classification. We include RGB, depth and skeletal data modalities in our experiments. For both feature extraction and dimension reduction, however, we use deep neural network based models. 

RNNs and their derivatives are also used frequently for sequence modeling in sign recognition tasks. One of the first researches in this domain can be found in \cite{murakami_gesture_1991}, where RNNs are used for sign recognition problem to classify signals obtained from a data glove. In this work, an RNN model is used for classifying the outputs of an Artificial Neural Network (ANN) model. The accuracy of this method is 98\% on their own data glove dataset that contains 10 signs. This shows the ability of RNNs to learn sequences. A more modern approach that are derived from RNNs are LSTMs. In \cite{tsironi_gesture_2016}, CNN and LSTM models are used together and compared to a CNN model. Differential images are generated using three frames and a combination of them with bitwise AND operation are fed to the proposed CNN-LSTM model. In the training and testing they used their own dataset which contains 9 gesture classes performed by 6 people. In total, the dataset contains 543 gesture samples in RGB format. Their CNN-LSTM model performed 91.67\% accuracy, whereas the classical CNN model has 77.78\% accuracy. \cite{nishida_multimodal_2016} investigates an architecture composed of LSTM layers to handle videos with variable-length gestures in order to capture the temporal correlation in gesture videos. To create a spatiotemporal representation, multiple temporal modalities are merged together, which produced an accuracy of 97.8\% on the SKIG dataset \cite{liu_learning_2013}, which contains 1080 samples that cover a total of 10 gestures. Another method for combining CNNs with LSTMs is proposed in \cite{nunez_convolutional_2018}. Human activity and hand gesture recognition problems are addressed using 3D data sequences obtained from full-body and hand skeletons, respectively. To this aim, they proposed a combination of a CNN and a LSTM recurrent network. Also, a two-stage training strategy which firstly focuses on CNN training and, secondly, adjusts the full method (CNN+LSTM), is presented. With this method, they presented a Jaccard index of 79.15 for Montalbano dataset. Similarly, we use CNNs for feature extraction and utilize an LSTM for classification of sequences in our research. We then compared the performance of CNN-LSTM model with our HMM approaches.

Focusing more on CNNs, \cite{pigou2014sign} uses them with the Montalbano dataset. Depth and RGB inputs are processed, combined and classified through two parallel streams. The inputs consist of two parts, upper body and hand. Attributes are extracted using the CNN model and the resulting feature matrices are given as input to the Fully Connected (FC) layer. With the constructed model, an accuracy of 91.7\% was achieved. \cite{pigou_beyond_2018} is a recent study with the Montalbano dataset that also takes the temporal aspect of data into account, pointing out the necessity of recurrence and showed that significant improvements are achieved when temporal convolutions are employed. They proposed an end-to-end trainable architecture that fuses the temporal convolutions and bidirectional recurrence. Additionally, they emphasize that using only a simple temporal feature pooling strategy is not enough to capture the temporal aspect of the video. In their findings, their proposed model, i.e. Temp Conv + RNN, achieves 93.76\% precision and 0.9 Jaccard index. In our preliminary work \cite{paper_siu}, we also used Montalbano dataset with CNN + LSTM based model with a Feature Pooling Module (FPM). A pretrained CNN model is used as feature extractor and FPM is appended at the end of the CNN model to encode features in multiple scales to be used with our LSTM model. RGB and depth modalities from Montalbano Dataset are used to evaluate this model and we achieved 93.15\% accuracy on isolated signs recognition task. 

There are also other approaches that achieve higher accuracies by optimizing their models on the task. \cite{neverova_moddrop_2016} deals with gesture detection and localization problem based on multi-scale and multi-modal deep learning. Depth and intensity data is used by focusing on the left and right hand. Additionally, Motion capture (Mocap) and Audio data are fused at several spatial and temporal scales leading to a significant increase in recognition rates. Using their proposed ModDrop training technique the robustness of the model is increased. With this method, a Jaccard index of 0.881 is reached for Montalbano Dataset \cite{montalbano_old}. In addition to ModDrop, the well-known Dropout method is used in \cite{li_modout_2017}. Dropout treats all units, visible or hidden, in the same way, thus ignoring any a priori information related to grouping or structure. They proposed Modout method, which is based on stochastic regularizations, is particularly useful in the multi-modal setting. It is capable of learning to decide if two modalities should be fused in a layer or not. The best results are achieved when Modout is used together with Dropout, resulting in a classification accuracy of 93.8\% and a Jaccard index of 0.888  for Montalbano Dataset. In the most recent research \cite{santos_dynamic_2020}, a classifier is designed using two Resnet CNNs, a soft-attention ensemble, and a fully connected layer. In their experiments, they report 94.58\% accuracy with Montalbano dataset. This result is the state-of-the-art on this dataset, using only the color information.

\section{Materials and Methods}

\subsection{Dataset and Preprocessing}
\label{s_dataset}
\begin{figure}[!h]
    \centering
    \begin{subfigure}{0.22\textwidth}
        \centering
        \includegraphics[scale=0.3,trim={240px 95px 190px 80px},clip]{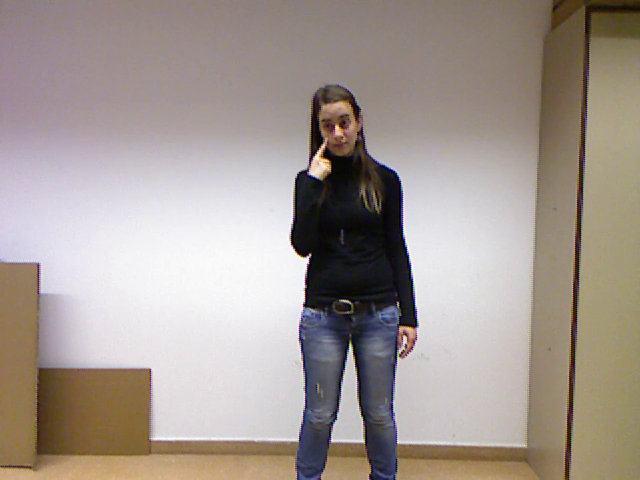}
        \caption{RGB image}
    \end{subfigure}\hfill%
    \begin{subfigure}{0.22\textwidth}
        \centering
        \includegraphics[scale=0.3,trim={240px 95px 190px 80px},clip]{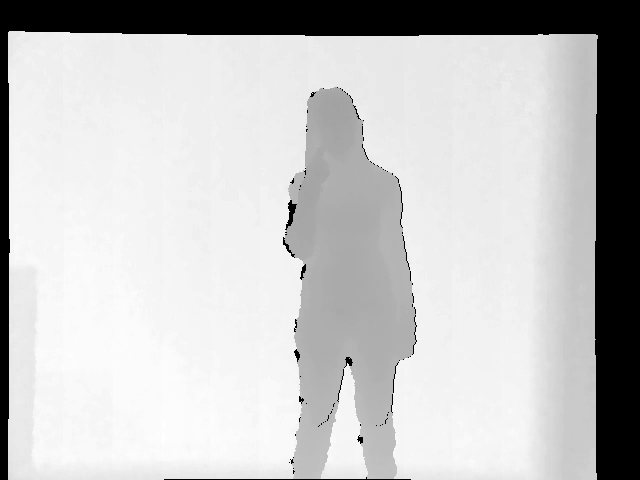}
        \caption{Depth image}
    \end{subfigure}\hfill%
    \begin{subfigure}{0.22\textwidth}
        \centering
        \includegraphics[scale=0.3,trim={240px 95px 190px 80px},clip]{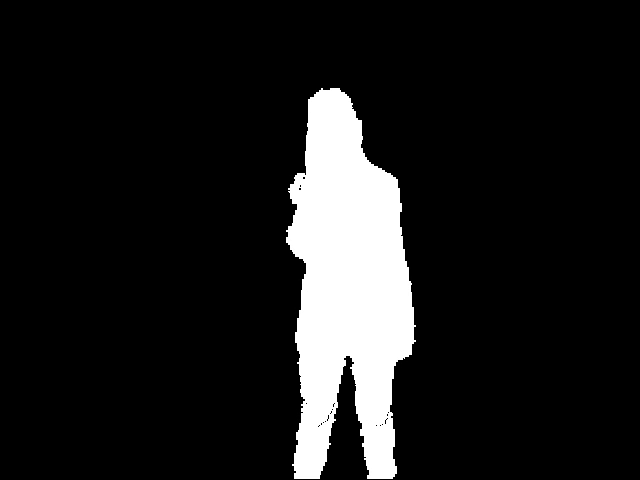}
        \caption{User mask}
    \end{subfigure}\hfill%
    \begin{subfigure}{0.22\textwidth}
        \centering
        \includegraphics[scale=0.3,trim={240px 95px 190px 80px},clip]{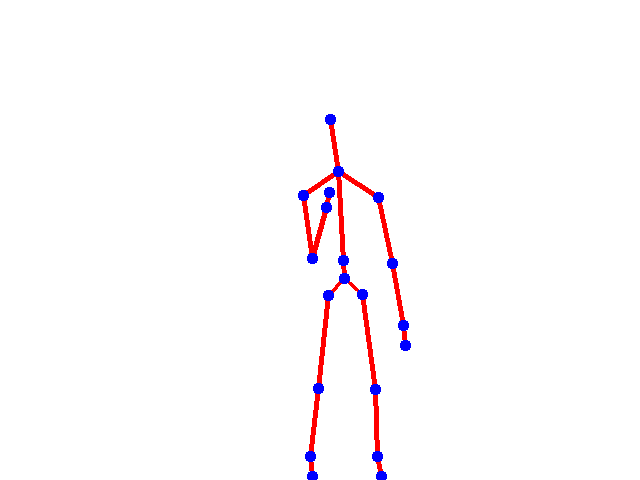}
        \caption{Skeletal data}
    \end{subfigure}                
    \caption{Sample images from Montalbano dataset.}
    \label{fig:dataset}
\end{figure}

In this research, we worked with the Montalbano Gesture dataset. This dataset contains 20 different Italian gestures that are performed by 27 different signers. The samples have various challenges in user environment, clothing, environment lighting and gesture complexity. The dataset is generated with Microsoft Kinect and has four different modalities: RGB, Depth, User Mask, and Skeletal data. Each sample video is recorded with 20 frames per second and a resolution of 640x480 pixels. A sample image is provided for different modalities from this dataset in Figure \ref{fig:dataset}. 

We observed that some of the User Mask data are too noisy and needs to be excluded from the final sample pool. After discarding the unusable data, we got 6787 training, 3428 validation and 3555 test samples. In total 13770 samples are used, each with varying number of frames.

\subsubsection{Preprocessing}
\label{s_proc}

\begin{figure}[!h]
    \begin{subfigure}{0.3\textwidth}
        \begin{center}
        \includegraphics[scale=0.2]{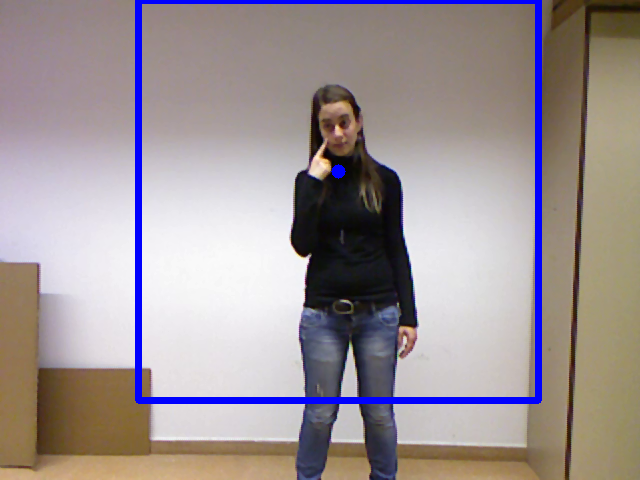}
        \caption{Cropping}
        \label{fig:cropping}
        \end{center}
    \end{subfigure}\hfill%
    \begin{subfigure}{0.3\textwidth}
        \begin{center}        
        \includegraphics[scale=0.22]{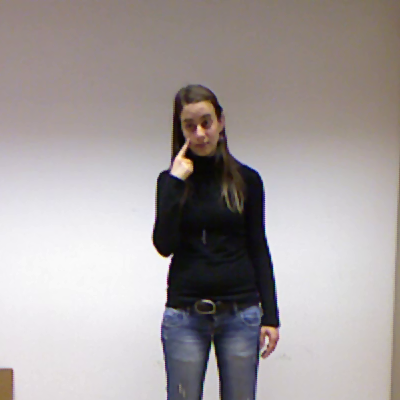}
        \caption{Preprocessed RGB}
        \end{center}
    \end{subfigure}\hfill%
    \begin{subfigure}{0.3\textwidth}        
        \begin{center}        
        \includegraphics[scale=0.22]{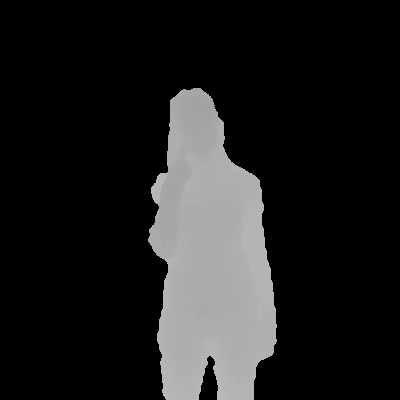}
        \caption{Preprocessed Depth}
        \label{fig:depth}
        \end{center}
    \end{subfigure}                
    \caption{Data Prepocessing}
    \label{fig:preprocess}
\end{figure}

In each video frame, in Montalbano dataset, there is a large redundant background data in addition to the signer. In order to focus more on the signer and reduce the redundant data, we cropped each frame to 400x400 pixels size. Cropping is performed easily by utilizing the skeletal data; x-coordinate of the shoulder center point is used to determine the center of the cropping square. Since the essential information in a sign video appears on the upper body of a signer, the cropping square is aligned to the top of the original RGB and depth images. A sample cropping is depicted in Figure \ref{fig:cropping}. The User Mask data from the dataset is used to reduce redundant information in the background of the depth images (Figure \ref{fig:depth}). Following the cropping operation, median filtering is applied to both RGB and depth images to reduce noise. Examples of resultant images can be seen in Figure \ref{fig:preprocess}. In addition to the RGB and depth modalities, we also utilized skeletal data. Skeletal data contains 3D spatial coordinates of 6 joints of the signer body. The points represent the hand, wrist and elbow joints for the left and right arm. We normalized the data from these joints by subtracting the coordinates of the shoulder center point and dividing the resulting point to the Euclidean distance between the shoulder center point and the hip center point. In addition, we reduced the data points in a frame by excluding all points below the waistline, resulting in an 18 dimensional vector.

%

\subsection{The Method}
\label{s_method}

Although deep features that are learned from CNN models can directly be used with deep sequence based models (i.e. LSTMs), they also suffer from higher dimensional data for complex tasks, such as sign recognition, when the number of samples for training is limited. The situation is worse for HMMs; using deep features directly without dimension reduction creates convergence problems during model training, i.e. curse of dimensionality problem. In this research, our primary purpose is to determine the components of an effective method to make HMM models work with deep features in this domain. There are a couple of reasons for this: (i) HMM models are relatively smaller than deep recurrent models and easy to train if the features are robust, (ii) Learning features from data is very advantageous since hand-crafted feature design is a challenging problem, (iii) Running HMMs is fast and easy in commodity computers compared to deep sequence models. 

\begin{figure}[!h]
      \begin{center}
        \includegraphics[width=\textwidth, height=9cm]{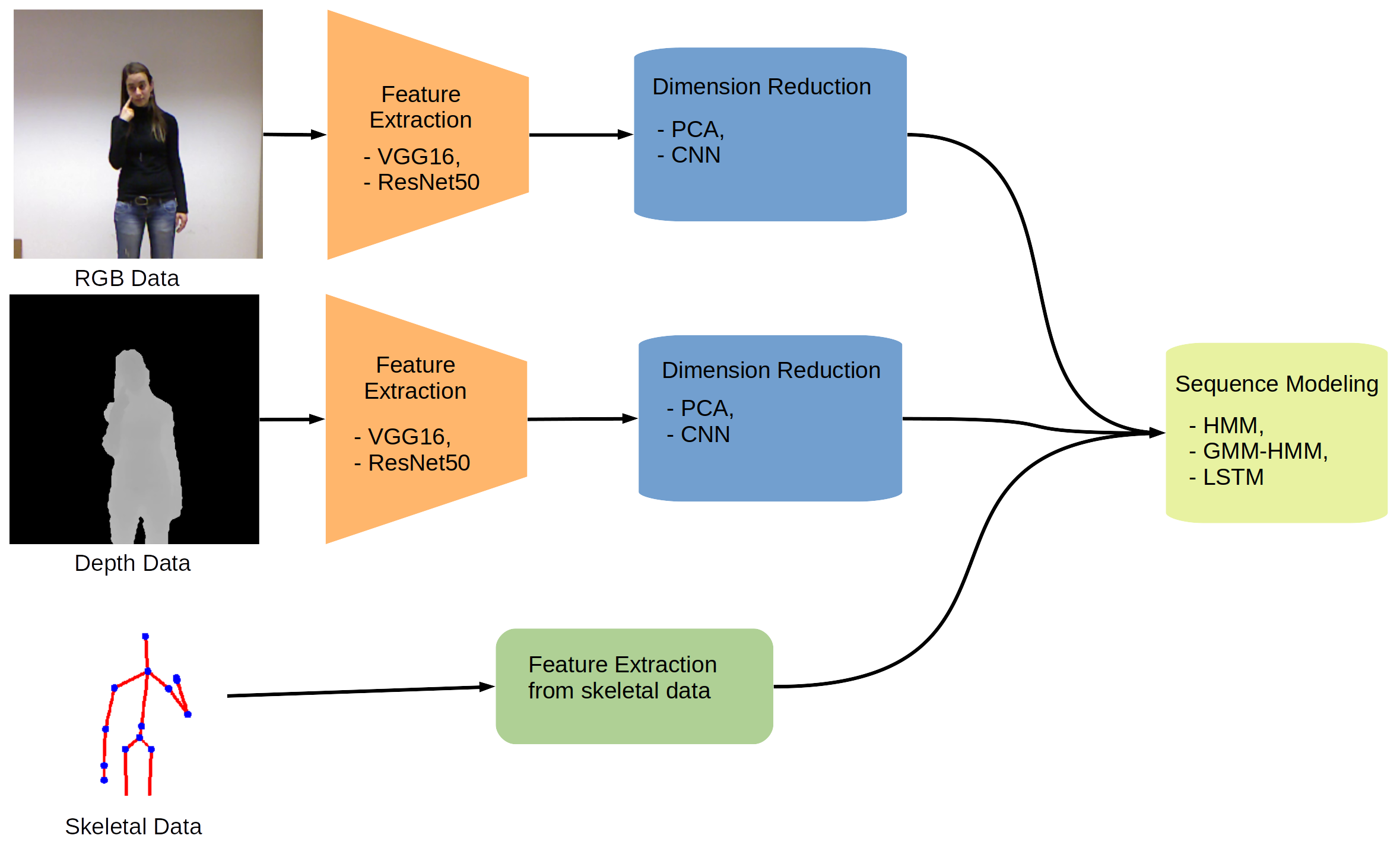}
        \caption{Proposed Framework.}
        \label{fig:model_des}
    \end{center} 
\end{figure}

Considering these issues, we created a framework that has three main modules: (1) Feature extraction module, (2) Dimension reduction module, (3) Sequence modelling module. Both RGB and depth data modalities are utilized through these three module structure in the given order to create variety of models with different settings in each module. As for the skeletal data, we only have a feature extraction module since the number of dimensions of the selected features are already small. For the skeletal data, our features are the selected joint coordinates that are normalized as we explained in Section \ref{s_proc}. A general overview of our framework is depicted in Figure \ref{fig:model_des}. As can be seen from the Figure, there are two parallel paths for the RGB and depth data modalities with similar model structure. Each path is trained separately with different inputs, using RGB and depth data. 

In the feature extraction module, we use two well-known pre-trained CNN network models, i.e. VGG16 and ResNet50 models. In the dimension reduction module, we evaluate PCA method, and two different CNN models that we designed and trained specifically for this task. In the sequence modelling module, we evaluate three models; HMMs with Gaussian emission probabilities, HMMs with Gaussian Mixtures in their emission probabilities (GMM-HMM) and LSTM models. Our strategy is first determining the best alternative models for isolated sign classification in the first two modules, i.e. feature extraction and dimension reduction modules, after a series of carefully designed experiments. Then we evaluate the performances of two of our HMM based sequence models with LSTM models under the same terms using the same deep features obtained at the end of the second module. We provide the details of each module in the proposed framework in the following subsections.

%
\subsubsection{Feature Extraction Module}
\label{s_method_1}

VGG16 \cite{vgg} and ResNet50 \cite{resnet} are both convolutional neural networks that have proven to be capable of extracting effective features from natural images on object recognition tasks. In this research, we utilize the pre-trained models of these two networks, which are trained using ImageNet dataset \cite{imagenet}, for feature extraction. In this scope, we did not re-train or fine-tune these models to adapt to Montalbano dataset; instead, we utilized both networks, from input through the outputs of the last convolutional layers as constant feature extractors. Our purpose in this research is to analyse and maximize the HMM model performances, considering the performance of our baseline LSTM model; not presenting the top accuracy score for a a particular dataset. Therefore, we believe that a standard deep feature extraction suffices our purpose.

VGG16 architecture contains 16 layers, where cascaded convolutional layers extract features from the input images. In this model, repeated application of small-sized convolutions are shown to be effective to generate sharp and better feature representations in many classification tasks. Application of small filter sizes in convolutions is also more efficient computationally. Following some of the convolutional layers, spatial pooling operation is carried out to increase the receptive field of the following convolutional filters. This structure enables encoding low-level features in the initial layers of the network, mid-level features in the middle and high-level features close to the end of the network. For this purpose, five max-pooling layers are used.

ResNet50, which is a derivation of the ResNet architecture, is also a deep convolutional neural network where residual connections are used to effectively increase the depth, hence the capacity, of the networks. Normally, when neural networks get deeper, it is more difficult to train them. Residual networks provide a solution to this problem with residual blocks. A residual block contains a branch leading out to a series of transformations F, whose outputs are added to the input x of the block:
\begin{equation}
\label{eq_res}
o = Activation(x + F(x))
\end{equation}
Designing layers as in (\ref{eq_res}), gradients are able to reach much deeper into the network without diminishing, during backpropagation. Contrary to VGG16, in ResNet50 pooling operation is not used between the convolutions, except once after the first convolutional layer. An average pooling layer is added right before the final fully connected layer.

%
\subsubsection{Dimension Reduction Module}
\label{s_method_2}

The outputs of the both architectures in the feature extraction module, are taken from the last convolutional layers of the models. Therefore, the dimensions of the feature vectors are large. In order to train sequence models with a limited dataset, without facing curse of dimensionality problem, we need to reduce the number of dimensions. For this purpose, we propose two different sets of CNN architectures. We generated a baseline model using the LSTM model that we proposed in our preliminary research \cite{paper_eurocon} in order to assess the performances of the proposed dimension reduction models (Section \ref{s_results}). 

The outputs of the last convolutional layers of VGG16 and ResNet50 models are 3D tensors, i.e. 512x12x12 and 2048x13x13, respectively; each containing a set of 2D tensors that are spatial responses of the filters in the last convolutional layer with different sizes. We first reduce each spatial filter response to a scalar value using one of the two selected pooling techniques, i.e. global average pooling (GAP) and global max pooling (GMP) methods. GAP reduces the filter responses of each filter by taking the averages of their spatial responses. GMP, on the other hand, selects the highest spatial response of each filter. After these pooling operations, 3D tensors are reduced to 1D tensors, i.e. to vectors. As a result, the output of ResNet50 model, is reduced to a 2048 dimensional vector, and similarly, the output of VGG16 is reduced to a vector of 512 dimensions. Still, these feature vector sizes are big for HMM training in this domain. Therefore, as to reduce the dimensions more, we applied PCA to reduce both network outputs to a vector of size 64, as our first method. 

\begin{figure}[!h]
\begin{center}
    \begin{subfigure}{0.30\textwidth}
        \begin{center}
        \includegraphics[width=0.65\textwidth]{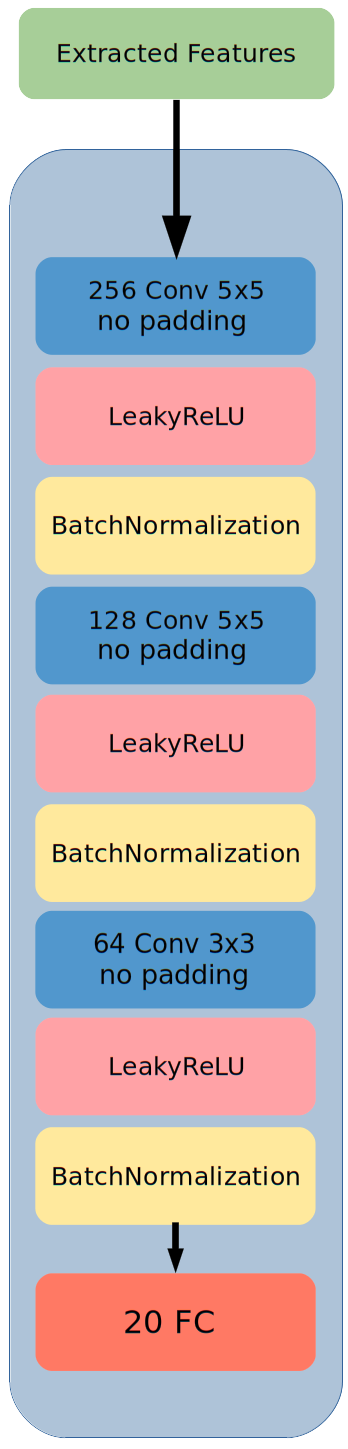}
        \caption{DR-CNN1 Architecture}
        \label{fig:feat_dim_v1}
        \end{center}
    \end{subfigure} \hfill    
    \begin{subfigure}{0.30\textwidth}
        \begin{center}        
        \includegraphics[width=0.63\textwidth]{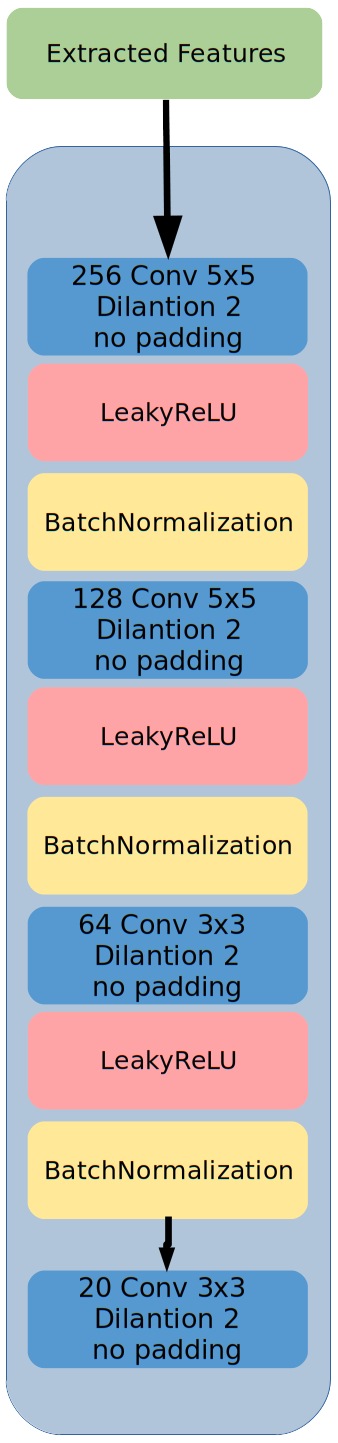}
        \caption{DR-CNN2a Architecture}
        \label{fig:feat_dim_v2a}
        \end{center}
    \end{subfigure}  \hfill
    \begin{subfigure}{0.30\textwidth}
        \begin{center}        
        \includegraphics[width=0.65\textwidth]{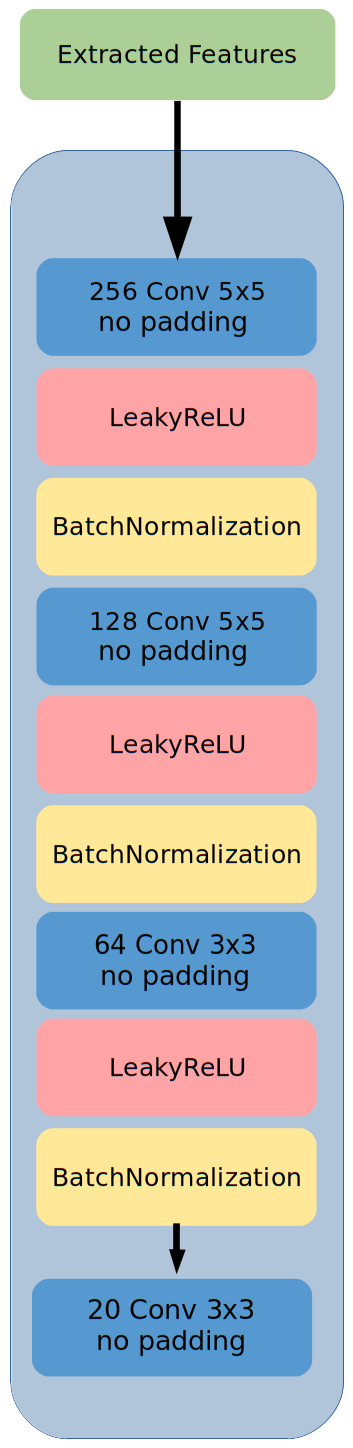}
        \caption{DR-CNN2b Architecture}
        \label{fig:feat_dim_v2b}
        \end{center}
    \end{subfigure} 
    
\end{center}
\caption{Dimension Reduction Networks.}
\end{figure}

Alternatively, we designed two CNN models for each network that are trained to learn a non-linear projection function from higher dimensional space to lower dimensional space from data. In the first one, a cascade of 2D convolutions are ended with a Fully Connected layer of size 20. Hence, the first architectures for the dimension reduction for ResNet50 and VGG16 models are the same (Figure \ref{fig:feat_dim_v1}). In the second architecture, we want to evaluate the performance of convolutions without a fully connected layer in the dimension reduction. Since the output sizes of the ResNet50 and VGG16 models are different, the second architectures needed to be designed separately for each network (Figure \ref{fig:feat_dim_v2a} and Figure \ref{fig:feat_dim_v2b}). In order to provide a larger field of view in the VGG16 output, the last max-pooling layer of the VGG16 model is removed in this setting; hence, the output tensor sizes for VGG16 became 512x25x25. To reduce the spatial dimension to 1 faster in VGG16 network we used dilated convolutions (Figure \ref{fig:feat_dim_v2a}). ResNet50 outputs are reduced without dilations (Figure \ref{fig:feat_dim_v2b}). We will refer to these three architectures as DR-CNN1 and DR-CNN2a and DR-CNN2b shortly from this on. For both of the feature extraction networks, each dimension reduction model is trained and evaluated separately, \textit{using only the training splits} of the selected dataset.

We train separate dimension reduction networks for depth and RGB modalities, for both networks, i.e. VGG16 and ResNet50. Since these networks are composed of CNN layers, in training, the parameter learning is performed using an optimization function on single frame image classification task. This is done by propagating the ground-truth video labels to each of their frames. We then evaluated performances of these models with different sequence models. Further details about training can be found in Section \ref{s_training}. Both feature extraction networks are paired with both dimension reduction networks, resulting in four possible combinations for each modality. The resultant features are used to train the LSTM, Gaussian HMM, GMM-HMM models separately, for each model and modality combination. We therefore completed substantial number of trainings to obtain the results we summarized in Section \ref{s_results}.

%

\subsubsection{Sequence Modelling Module}
\label{s_seq}
Hidden Markov Models \cite{HMMIntro} are conventional methods in temporal pattern recognition that have been used widely for  sequence modelling in various domains such as action recognition, speech recognition, handwriting recognition and sign language recognition. Although LSTM models have been lately used more frequently in sequence modelling in the domains with a lot of data, HMMs are strong competitors to LSTM models, especially when the data is limited.

\begin{figure}[!h]
      \begin{center}
        \includegraphics[width=0.6\textwidth]{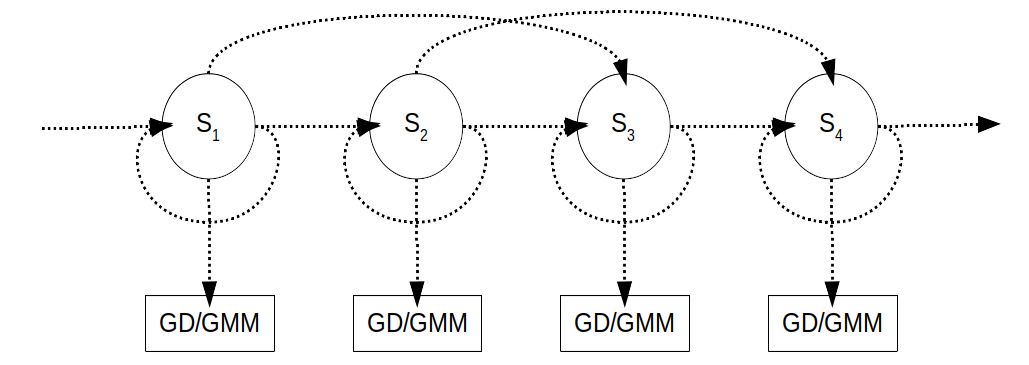}
        \caption{A four-state HMM with Gaussain Distribution and Gaussian Mixture Model.}
        \label{fig:hmm_model}
    \end{center} 
\end{figure}

The parameter set of an N-state HMM consists of an initial state distribution vector for each state, a state transition probability matrix, i.e. of size NxN, and the parameter set of its output probability distributions, assuming that it is represented in a parametric form. A simple diagram of a 4-state HMM model is depicted in Figure \ref{fig:hmm_model}. In this research, since the observed samples are features with multiple real values, we worked with two variants of Gaussian distributions as the output probability distribution of each state: (1) a multivariate Gaussian Distribution, (2) a Gaussian Mixture Model. We will refer to the HMMs with Gaussian emission probabilities shortly as HMMs , and HMMs with Gaussian Mixture emissions shortly as GMM-HMMs from now on.

HMMs are called ergodic or left-to-right HMM models depending on the type of transitions allowed among the states of the model. HMMs that allow transition from an emitting state to any other state in the model are called ergodic HMMs. This type of transition allows modelling cyclic patterns, like hand waving action in a video. Left-to-right HMMs allow transition from any emitting state to only the successor of that state or the state itself. This configuration is more appropriate for modelling acyclic sequences that have some trajectory from beginning to end. In this work, we experimented with both transition models to determine the best alternative in this problem.

Convergence during the training of an HMM model is heavily affected  by the curse of dimensionality problem \cite{CurseOfDimProblem}. To minimize this effect, a robust set of features with a reasonable number of dimensions, which is usually determined empirically depending on the training data, need be extracted from the frames of a sign language video. In this context, we first extract the features of the RGB data frames using one of the two well-known pretrained CNN models. Then the obtained features are passed as an input to our dimension reduction network, which reduces the feature size to 20. As a result, a 20 dimensional feature vector is produced for each RGB frame in a video. The same process is applied to the depth data, independently from the RGB data. Thus, each depth video frame in a sample is also reduced to a 20 dimensional vector. As for the Skeletal data, 18 dimensional feature vectors are extracted as it is described in Section \ref{s_proc}.

In classifications, for a given sequence, HMMs provide the likelihood of the model for generating that sequence. In this setting, for Montalbano dataset, there are 20 signs, hence we train 20 different HMM models for each sign. In order to classify a given sequence, the likelihoods of all the trained HMM sign models are computed; the category of the sign is the respective sign of the model that has the highest likelihood score among all the models.

For the LSTM model, we utilized the architecture that we proposed in our preliminary research; the architectural details of the model and training can be found in \cite{paper_eurocon}. Minor changes are applied to that model implementation to make it work with varying number of frames, instead of using sequences with a fixed frame length. 

\subsubsection{Training and Implementation Details}
\label{s_training}
Implementation of the deep neural networks are done using PyTorch framework \cite{pytorch}. For the implementation of HMMs, Pomegranate framework \cite{pomegranate} is used because of its numerical stability and ease of use.

The dimension reduction networks are trained with Adam optimizer \cite{adamopt}. Adam’s beta1 and beta2 parameters are set to 0.9 and 0.999, respectively. During training, the learning rate is set to 0.001. Cross entropy loss \cite{cross} is used for error propagation for each image, with their labels taken from the sign label of the respective video sample. The batch size is set as 32. The networks converge after three epochs, showing no further improvement afterwards.

The LSTM models are also trained with the Adam optimizer. Adam’s parameters are used with the same values as they are during the training of the dimension reduction networks. During training, the learning rate is set to 0.001. Cross entropy loss is used for error propagation on whole sign sequences. The the hidden layer size is tested with 64, 128 and 256. When dimension reduction is skipped and the output of the feature extractor is fed into the LSTM directly, 256 hidden units performed better than the rest. However, when dimension reduction is applied, 64 hidden units perform the best. To determine the best batch size, it is set to different values, i.e. 8, 16, 32 and 64. Best results are achieved with 64 and the presented results are based on this setting.

When training HMMs we started with 18 dimensional Skeletal Data, i.e. we did not use RGB and Depth modalities at first; since the numerical instability increases with the increase in the number of dimensions. We tested Left-to-Right and Ergodic models and concluded that Ergodic models perform higher compared to Left-to-Right models. We continue with Ergodic models for the rest of the research. We used Viterbi algorithm \cite{viterbi} while training HMMs and generating a likelihood score of a sequence. When using multiple modalities, we apply two approaches to merge the results. We call the first approach \textit{Max Merge} approach. The Max Merge approach determines the label of the sample considering the label of the HMM that gives the highest likelihood among the modalities that are chosen for the given experiment i.e. RGB, Depth, Skeletal Data. The second approach is the \textit{Concat} approach, which concatenates the feature vectors of all the modalities that we want to use for the experiment and trains a single HMM per concatenated feature vectors. The label of the single highest likelihood is chosen as the correct label for a given sample.

\section{Results and Discussions}
\label{s_results}

In our preliminary work, we obtained the best classification accuracy using the concatenated RGB and depth modalities, with fixed frame rate using Global Max Pooling after the feature extraction \cite{paper_eurocon}. Fixed frame rate is implemented to represent each sign video with 40 frames. The best result, which is 93.19\% accuracy, is obtained using Resnet50 model features. When we changed the fixed frame rate to variable frame rate, we obtained the results in Table \ref{tab:lstm_GAPGMP_RGBDepth}. The best result is obtained again with GMP using ResNet50 model, yet the performance of the LSTM model is slightly lower than the fixed frame rate implementation.

\begin{table}[H]
\centering
\begin{tabular}{ccc}
\toprule
\tabhead{Pooling Method}   & \tabhead{VGG-16 Model} & \tabhead{Resnet50 Model} \\ \midrule
GAP           & 89.75\%      & 90.90\%  \\ \hline
GMP           & 88.07\%      & \textbf{91.07\%} \\
\bottomrule
\end{tabular}
\caption{LSTM model performances without dimension reduction, with varying frame lengths for RGB+Depth modalities.}
\label{tab:lstm_GAPGMP_RGBDepth}
\end{table}

When we use the same features for HMM model training, i.e. without dimension reduction, we observe that HMM models can not converge properly with these feature dimensions. It is the reason that we integrated the second module in our framework, i.e. dimension reduction module, to train HMMs on this problem to reduce convergence problems. 

\paragraph{\textit{LSTM Model Performances After Dimension Reduction:}} 

In order to evaluate the second module, i.e. the dimension reduction models, in our framework, we designed a set of experiments using again our preliminary LSTM model. These models will be serving as baselines for assessing the HMM performances.

\begin{table}[H]
\centering
\begin{tabular}{cccc}
\toprule
\tabhead{Dimension Reduction Method}   & \tabhead{Feature Size}	&\tabhead{VGG-16 Model} & \tabhead{Resnet-50 Model} \\ \midrule
GAP+PCA       	& 64	&78.71\%       & 75.97\%  \\ \hline
GMP+PCA      	& 64	&73.43\%      & 55.28\% \\ \hline
DR-CNN1			& 20	&85.25\%      & \textbf{86.06\%} \\ \hline
DR-CNN2a		& 20	&81.74\%      & - \\ \hline
DR-CNN2b		& 20	&-      	& 85.46\% \\
\bottomrule
\end{tabular}
\caption{LSTM model performances after dimension reduction, with varying length frames for RGB+Depth modalities. The hidden layer size of the LSTM models are 64.}

\label{tab:lstm_DimReduction}
\end{table}

As shown in Table \ref{tab:lstm_DimReduction}, the feature dimensions are reduced to 20 with the DR-CNN models and to 64 with the PCA method. Reducing the feature dimension for less than 64, reduces the classification accuracies for the PCA model. For varying frame rates and RGB+Depth modalities, the best result is obtained with ResNet50 model for feature extraction and using our first CNN model, i.e. DR-CNN1, in the dimension reduction. Resnet50 features are reduced from 2048 to 20. Although the feature size is reduced substantially, to around 1\% of the original feature size, the classification accuracy is still 86.06\%; only less than 5\% reduction is observed compared to the best score in Table \ref{tab:lstm_GAPGMP_RGBDepth}.

\paragraph{\textit{Experiments with HMM Models:}}

We started to work with HMM models using Skeletal features, since the number of dimensions, i.e. 18, is considerably smaller than the deep features that are obtained with RGB and depth modalities. Our purpose is to determine some of the hyperparameters of the HMM models before dealing with dimension reduction. Our initial experiments with HMMs and Skeletal data revealed that ergodic HMMs yielded better results than left-to-right HMMs, as can be seen in Table \ref{tab:hmm_skel}. The data also shows that increasing the number of states in an HMM does not always give better results; the model performances start to decrease slightly with more than 20 states.

\begin{table}[H]
\centering
\begin{tabular}{ccc}
\toprule
\tabhead{States}   & \tabhead{Left-to-Right} & \tabhead{Ergodic} \\ \midrule
10       & 74.26\%       & 75.78\% \\ \hline
15       & 76.37\%       & 78.90\% \\ \hline
20       & 72.74\%       & \textbf{79.41\%} \\ \hline
25       & -             & 78.87\% \\
\bottomrule
\end{tabular}
\caption{HMM Results with Skeletal Data. Best score is depicted using bold font.}
\label{tab:hmm_skel}
\end{table}

For RGB and depth modalities, we first evaluated HMM model performances using PCA dimension reduction method, reducing the feature vectors to 64 dimensions for each modality. Due to numerical instability issues during training, using larger features (64 and 128 when RGB and depth are combined, instead of 18), we observed decline in the performances when we design HMMs with more than 10 states. Therefore, the hyperparameters with RGB and depth modalities are set as ergodic HMM models with 10 states in the following experiments. Table \ref{tab:hmm_pca} shows the results with this configuration for both VGG16 and ResNet50 features. The best score is obtained with PCA method with Resnet50 model, i.e. 63.35\%, combining both RGB and depth modalities with \textit{Max Merge} method.

\begin{table}[H]
\centering
\begin{tabular}{ccc}
\toprule
\tabhead{Modalities} & \tabhead{VGG16} & \tabhead{ResNet50}  \\ \midrule
RGB     & 57.38\%         & 63.04\%        \\ \hline
Depth     & 58.20\%         & 57.83\%        \\ \hline
\begin{tabular}[c]{@{}c@{}}RGB + Depth\\ (Max Merge)\end{tabular} & 58.34\%         & \textbf{63.35\%} \\
\bottomrule
\end{tabular}
\caption{10 State HMM Results with PCA Dimension Reduction (64 Dimensions).}
\label{tab:hmm_pca}
\end{table}

In order to assess the performances of our dimension reduction networks, i.e. DR-CNN1 and DR-CNN2(a/b), we performed a set of experiments using the ergodic HMM models. The number of HMM states in these experiments is set to 10 for both HMMs and GMM-HMMs, while the number of mixtures was set to 3 for all GMM-HMMs for the preliminary analysis to determine the effective models in the first two modules in our pipeline. The test results are depicted in Table \ref{tab:comp_table}. 

\begin{table}[H]
\centering
\begin{adjustbox}{width=1\textwidth}
\begin{tabular}{|c|c|c|c|c|c|c|c|c|c|}
\hline
\multirow{2}{*}{\textbf{Preprocessing}} & \multirow{2}{*}{}        & \multicolumn{4}{c|}{\textbf{VGG16}}                            & \multicolumn{4}{c|}{\textbf{ResNet50}}                         \\ \cline{3-10} 
                               &                          & \multicolumn{2}{c|}{\textbf{DR-CNN1}} & \multicolumn{2}{c|}{\textbf{DR-CNN2a}} & \multicolumn{2}{c|}{\textbf{DR-CNN1}} & \multicolumn{2}{c|}{\textbf{DR-CNN2b}} \\ \hline
\textbf{Sequence Model}        &                          & GMM-HMM     & HMM         & GMM-HMM     & HMM         & GMM-HMM     & HMM         & GMM-HMM     & HMM         \\ \hline
\multirow{5}{*}{\textbf{Modalities}}    & RGB                      & 85.99\%     & 85.06\%     & 81.07\%     & 80.90\%     & 87.48\%     & 86.89\%     & 86.95\%     & 86.19\%     \\ \cline{2-10} 
                               & Depth                    & 38.51\%     & 32.77\%     & 40.51\%     & 35.36\%     & 44.53\%     & 39.63\%     & 43.68\%     & 38.68\%     \\ \cline{2-10} 
                               & \begin{tabular}[c]{@{}c@{}}RGB + Depth\\ (Max Merge)\end{tabular} & 38.65\%     & 32.86\%     & 40.84\%     & 35.50\%     & 58.31\%     & 53.98\%     & 43.74\%     & 38.73\%     \\ \cline{2-10} 
                               & \begin{tabular}[c]{@{}c@{}}RGB + Depth\\ (Concat)\end{tabular}    &      -      &      -      &     -       &    -        & 86.98\%     & 87.59\%     &      -       &     -        \\ \cline{2-10} 
                               & RGB + Skeletal           &    -        &       -     &       -     &     -       & 86.02\%     & \textbf{88.52\%}     &    -       &     -       \\ \hline
\end{tabular}
\end{adjustbox}
\caption{General Comparison. HMMs have 10 states, GMM-HMMs have 10 states with 3 mixtures.}
\label{tab:comp_table}
\end{table}

These experiments enable us to compare model performances for different modalities using different dimension reduction networks. The table shows that using ResNet50 as the feature extractor yields better results than using VGG16 model, consistent with our previous experiments. Furthermore, DR-CNN1 reduction model results in higher accuracies than the DR-CNN2b model. These preliminary results show that the best feature extraction pipeline is comprised of a ResNet50 model in the first module and DR-CNN1 model in the second module for the dimension reduction. Hence, the following experiments are conducted with these configurations in feature extraction, focusing on the hyperparameter optimization for HMM model parameters.

In the first set of experiments, we worked with a changing state size and data modalities of Gaussian HMM models. The results of our experiments are depicted in Table \ref{tab:hmm_search}. After observing higher scores with RGB+Skeletal features, we expanded the experiments by reducing the number of states more after observing an increase in the performances in that direction. The best result is obtained concatenating RGB and Skeletal features with 6 state HMMs with Gaussian emissions.

\begin{table}[H]
\centering
\begin{adjustbox}{width=1\textwidth}
\begin{tabular}{|c|c|c|c|c|c|}
\hline
\textbf{Feature Extraction Model} & \multicolumn{5}{c|}{\textbf{Resnet50}}                                                            \\ \hline
\textbf{Dimension Reduction Model}       & \multicolumn{5}{c|}{\textbf{DR-CNN1}}                                                            \\ \hline
\textbf{HMM States}            & \textbf{RGB} & \textbf{Depth} & \textbf{\begin{tabular}[c]{@{}c@{}}RGB + Depth\\ (Max Merge)\end{tabular}} & \textbf{\begin{tabular}[c]{@{}c@{}}RGB + Depth\\ (Concat)\end{tabular}} & \textbf{\begin{tabular}[c]{@{}c@{}}RGB + Skeletal\\ (Concat)\end{tabular}} \\ \hline
\textbf{6}                   &    -         &  -             &  -                 &  -                 & \textbf{90.15\%} \\ \hline
\textbf{8}                   &    -         &  -             &  -                 &  -                 & 89.58\%                 \\ \hline
\textbf{10}                  & 86.89\%      & 39.63\%        & 53.98\%            & 87.59\%            & 88.52\%                 \\ \hline
\textbf{12}                  & 87.23\%      & 41.18\%        & 55.61\%            & 87.40\%            & 88.83\%                 \\ \hline
\textbf{15}                  & 87.37\%      & 42.28\%        & 56.74\%            & 87.59\%            & 88.38\%                 \\ \hline
\textbf{17}                  & 86.81\%      & 42.67\%        & 57.30\%            & 87.59\%            & 88.10\%                 \\ \hline
\textbf{20}                  & 86.89\%      & 44.16\%        & 58.48\%            & 87.65\%            & 87.79\%                 \\ \hline
\end{tabular}
\end{adjustbox}
\caption{HMM Model performances with varying state and data modalities, using Resnet50 and DR-CNN1 models.}
\label{tab:hmm_search}
\end{table}

We conducted similar experiments for GMM-HMMs with the addition of the number of mixtures parameter in our search space. The experiments are performed with varying data modalities similar to HMM experiments above. The test results are depicted in Table \ref{tab:gmmhmm_search}. A similar parameter search is performed for state-mixture pairs after observing higher scores with RGB+Skeletal feature concatenations. The best result is obtained using GMM-HMMs with 4 states and 3 mixtures.

\begin{table}[H]
\centering
\begin{adjustbox}{width=1\textwidth}
\begin{tabular}{|c|c|c|c|c|c|}
\hline
\textbf{Feature Extraction Model}                                               & \multicolumn{5}{c|}{\textbf{ResNet50}}                                                            \\ \hline
\textbf{Dimension Reduction Model}                                                     & \multicolumn{5}{c|}{\textbf{DR-CNN1}}                                                            \\ \hline
\textbf{\begin{tabular}[c]{@{}c@{}}GMM-HMM\\ States - Mixtures\end{tabular}} & \textbf{RGB} & \textbf{Depth} & \textbf{\begin{tabular}[c]{@{}c@{}}RGB + Depth\\ (Max Merge)\end{tabular}} & \textbf{\begin{tabular}[c]{@{}c@{}}RGB + Depth\\ (Concat)\end{tabular}} & \textbf{\begin{tabular}[c]{@{}c@{}}RGB + Skeletal\\ (Concat)\end{tabular}} \\ \hline
\textbf{2 - 3}                                                             &     -        &  -             & -                  & -                  & 88.21\%                 \\ \hline
\textbf{3 - 3}                                                             &     -        &  -             & -                  & -                  & 88.83\%                 \\ \hline
\textbf{3 - 5}                                                             &     -        &  -             & -                  & -                  & 88.35\%                 \\ \hline
\textbf{4 - 3}                                                             &     -        &  -             & -                  & -                  & \textbf{89.82\%}            \\ \hline
\textbf{5 - 3}                                                             & 86.84\%      & 42.73\%        & 59.96\%            & 87.54\%            & 88.80\%                 \\ \hline
\textbf{8 - 3}                                                             & 87.14\%      & 44.92\%        & 59.04\%            & 86.72\%            & 88.33\%                 \\ \hline
\textbf{8 - 5}                                                             & 87.06\%      & 45.15\%        & 59.07\%            & 87.20\%            & 86.67\%                 \\ \hline
\textbf{10 - 2}                                                            & 87.00\%      & 44.11\%        & 58.51\%            & 87.34\%            & 87.37\%                 \\ \hline
\textbf{10 - 3}                                                            & 87.48\%      & 44.53\%        & 58.31\%            & 86.98\%            & 86.02\%                 \\ \hline
\textbf{10 - 5}                                                            & 87.23\%      & 45.40\%        & 59.75\%            & 85.85\%            & -                       \\ \hline
\textbf{20 - 3}                                                            & 86.61\%      & 46.50\%        & 60.08\%            &  -                 &  -                      \\ \hline
\end{tabular}
\end{adjustbox}
\caption{GMM-HMM model performances with varying state and mixture parameters, using Resnet50 and DR-CNN1 models.}
\label{tab:gmmhmm_search}
\end{table}

\paragraph{\textit{Overall Performances of the Sequence Models:}}
The best overall results for different sequence models are summarized in Table \ref{tab:best}. Our preliminary LSTM model with reduced feature dimensions are set as the baseline of this evaluation; since we have already shown that this model performs comparable performance with Montalbano dataset \cite{paper_eurocon}. The best result with this baseline model is obtained using Resnet50 model and DR-CNN1 model, i.e. 86.80\%. 

The HMM model performances are summarized in the same table, with the LSTM baseline. As can be seen from the Table, PCA method can not exceed the baseline model; even the best performing PCA model performs quite poorly, with a 20\% margin, compared to the baseline. DR-CNN1 model outperforms the PCA model considerably. HMM and GMM-HMM performances with feature concatenations, i.e. RGB+Depth or RGB+Skeletal, exceed the baseline LSTM model. Although the best performances between HMM and GMM-HMM models are close to each-other, HMM performs slightly better, i.e 90.15\%.

In the initial experiments with HMMs, we observed convergence problems due to the high deep feature dimensions for HMM models. After we reduce the dimensions with our proposed CNN models, we observe that HMM models can even exceed LSTM model performances with the same set of features. Even, when we compare HMM model performance with our preliminary LSTM model, without dimension reduction, i.e. 91.07\%, the performance of the HMM model, which is 90.15\%, is still comparable with it. Considering the training and run-time benefits of HMMs, this research provides a framework that makes HMMs useful for the challenging sign video classification by using deep features effectively.

\begin{table}[H]
\centering
\begin{tabular}{lcclc}
\toprule
\tabhead{Model}                   & \tabhead{Modality}         & \tabhead{Test Accuracy} \\ \midrule

LSTM-Baseline (Resnet50+DR-CNN1+LSTM)   
								& RGB+Skeletal (Concat) & \textbf{86.80}\%       \\ \hline
Resnet50+PCA+HMM                 & RGB+Depth (Max Merge)           & 63.35\%     \\ \hline
\multirow{2}{*}{Resnet50+DR-CNN1+GMM-HMM} 
								& RGB                   & 87.48\%       \\
					              & RGB+Depth (Concat)           & 87.54\%       \\
                                  & RGB+Skeletal (Concat)        & \textbf{89.82}\%       \\ \hline

\multirow{3}{*}{Resnet50+DR-CNN1+HMM}    
								 & Skeletal              & 79.41\%       \\
                                  & RGB                   & 87.37\%       \\
                                  & RGB+Depth (Concat)           & 87.65\%       \\
                                  & RGB+Skeletal (Concat)        & \textbf{90.15\%}       \\

\bottomrule
\end{tabular}
\caption{Best results of the sequence models using different data modalities.}
\label{tab:best}
\end{table}

\section{Conclusion}
\label{s_conc}
In this research, we provided a three stage framework that enables isolated sign classification problem using both LSTM based and HMM based sequence models. The first module is used to extract features, the second module is used to reduce dimension, if necessary, and the third module serves as a sequence classifier. For feature extraction, a pretrained version of either VGG16 or ResNet50 models is used. Dimension reduction is handled with PCA or one of the two CNN architectures that we proposed in this research. For sequence classification LSTMs, Gaussian HMMs and GMM-HMMs are utilized.

During our research we observed that pretrained ResNet50 model performs better than VGG16 model as a feature extractor for our purpose. DR-CNN1 architecture, in conjunction with ResNet50, yields better results compared to our second CNN-based dimension reduction architecture or the PCA method. We obtained the best accuracy, i.e. 90.15\%, with a Gaussian HMM model using concatenated RGB and Skeletal data. Using the same deep features, our baseline LSTM model achieves its best accuracy, i.e. 86.80\%, using RGB and Depth data; however, when we train the same LSTM model \cite{paper_eurocon} without dimension reduction, we observe a slightly better accuracy, i.e. 91.07\%.

We observed and showed with empirical data that the second module is necessary for HMM models and helps to work with them without convergence problems. We experimented with various combinations of feature extraction, dimension reduction and recognition models. We provided a step by step analysis by changing the method in each module of the proposed framework, so as to obtain a comparable classification result using Gaussian HMMs and GMM-HMMs. We show that utilizing the same deep features, with feasible feature dimensions, HMMs can be an efficient alternative to LSTM models. They have fewer number of parameters than LSTMs, hence require fewer training samples. Moreover, their running time is considerably faster than deep sequence models. 

On the other hand, when the feature dimensions are large, HMMs suffer from \textit{the curse of dimensionality problem} more than LSTM models. LSTMs can handle large dimensions better when provided with enough samples. When the number of samples is limited, however, HMMs can be a better alternative; only if features are represented in a lower dimensional space effectively. In this work, we show that in addition to the feature extraction, dimension reduction can also be made more effectively using CNN models.

\section*{Acknowledgement}
The research presented is part of a project funded by The Scientific and Technological Research Council of Turkey (TÜBİTAK) under the grant number 217E022.

\%bibliographystyle{ieeetr}
\bibliographystyle{spmpsci}
\footnotesize \bibliography{refs}

\begin{thebibliography}{10}
\providecommand{\url}[1]{{#1}}
\providecommand{\urlprefix}{URL }
\expandafter\ifx\csname urlstyle\endcsname\relax
  \providecommand{\doi}[1]{DOI~\discretionary{}{}{}#1}\else
  \providecommand{\doi}{DOI~\discretionary{}{}{}\begingroup
  \urlstyle{rm}\Url}\fi

\bibitem{akram_visual_2012}
Akram, S., Beskow, J., Kjellstrom, H.: Visual {Recognition} of {Isolated}
  {Swedish} {Sign} {Language} {Signs}.
\newblock arXiv:1211.3901 [cs]  (2012).
\newblock \urlprefix\url{http://arxiv.org/abs/1211.3901}.
\newblock ArXiv: 1211.3901

\bibitem{cheok_review_2019}
Cheok, M.J., Omar, Z., Jaward, M.H.: A review of hand gesture and sign language
  recognition techniques.
\newblock International Journal of Machine Learning and Cybernetics
  \textbf{10}(1), 131--153 (2019).
\newblock \doi{10.1007/s13042-017-0705-5}.
\newblock \urlprefix\url{https://doi.org/10.1007/s13042-017-0705-5}

\bibitem{cooper_sign_2012}
Cooper, H., Ong, E.J., Pugeault, N., Bowden, R.: Sign {Language} {Recognition}
  using {Sub}-{Units}.
\newblock Journal of Machine Learning Research \textbf{13}(Jul), 2205--2231
  (2012).
\newblock \urlprefix\url{http://www.jmlr.org/papers/v13/cooper12a.html}

\bibitem{hog_method}
Dalal, N., Triggs, B.: Histograms of oriented gradients for human detection.
\newblock In: 2005 {IEEE} {Computer} {Society} {Conference} on {Computer}
  {Vision} and {Pattern} {Recognition} ({CVPR}'05), vol.~1, pp. 886--893 vol. 1
  (2005).
\newblock \doi{10.1109/CVPR.2005.177}.
\newblock ISSN: 1063-6919

\bibitem{escalera_challenges_2017}
Escalera, S., Athitsos, V., Guyon, I.: Challenges in {Multi}-modal {Gesture}
  {Recognition}.
\newblock In: S.~Escalera, I.~Guyon, V.~Athitsos (eds.) Gesture {Recognition},
  The {Springer} {Series} on {Challenges} in {Machine} {Learning}, pp. 1--60.
  Springer International Publishing, Cham (2017).
\newblock \doi{10.1007/978-3-319-57021-1_1}.
\newblock \urlprefix\url{https://doi.org/10.1007/978-3-319-57021-1_1}

\bibitem{motalbano}
Escalera, S., Bar{\'o}, X., Gonzalez, J., Bautista, M.A., Madadi, M., Reyes,
  M., Ponce-L{\'o}pez, V., Escalante, H.J., Shotton, J., Guyon, I.: Chalearn
  looking at people challenge 2014: Dataset and results.
\newblock In: Workshop at the European Conference on Computer Vision, pp.
  459--473. Springer (2014)

\bibitem{montalbano_old}
Escalera, S., Gonzàlez, J., Baró, X., Reyes, M., Lopes, O., Guyon, I.,
  Athitsos, V., Escalante, H.: Multi-modal gesture recognition challenge 2013:
  dataset and results.
\newblock In: Proceedings of the 15th {ACM} on {International} conference on
  multimodal interaction, {ICMI} '13, pp. 445--452. Association for Computing
  Machinery, Sydney, Australia (2013).
\newblock \doi{10.1145/2522848.2532595}.
\newblock \urlprefix\url{https://doi.org/10.1145/2522848.2532595}

\bibitem{viterbi}
Forney, G.: The viterbi algorithm.
\newblock Proceedings of the IEEE \textbf{61}(3), 268--278 (1973).
\newblock \doi{10.1109/PROC.1973.9030}.
\newblock Conference Name: Proceedings of the IEEE

\bibitem{grobel_isolated_1997}
Grobel, K., Assan, M.: Isolated sign language recognition using hidden {Markov}
  models.
\newblock In: Computational {Cybernetics} and {Simulation} 1997 {IEEE}
  {International} {Conference} on {Systems}, {Man}, and {Cybernetics}, vol.~1,
  pp. 162--167 vol.1 (1997).
\newblock \doi{10.1109/ICSMC.1997.625742}

\bibitem{resnet}
He, K., Zhang, X., Ren, S., Sun, J.: Deep residual learning for image
  recognition.
\newblock In: Proceedings of the IEEE conference on computer vision and pattern
  recognition, pp. 770--778 (2016)

\bibitem{jan_discriminative_2018}
{Jan Hendrik Combrink}: Discriminative training of hidden {Markov} {Models} for
  gesture recognition.
\newblock Master's thesis, University of Cape Town (2018).
\newblock \urlprefix\url{https://open.uct.ac.za/handle/11427/29267}

\bibitem{jie_huang_sign_2015}
{Jie Huang}, {Wengang Zhou}, {Houqiang Li}, {Weiping Li}: Sign {Language}
  {Recognition} using {3D} convolutional neural networks.
\newblock In: 2015 {IEEE} {International} {Conference} on {Multimedia} and
  {Expo} ({ICME}), pp. 1--6 (2015).
\newblock \doi{10.1109/ICME.2015.7177428}

\bibitem{CurseOfDimProblem}
Keogh, E., Mueen, A.: Curse of Dimensionality, pp. 314--315.
\newblock Springer US, Boston, MA (2017).
\newblock \doi{10.1007/978-1-4899-7687-1_192}.
\newblock \urlprefix\url{https://doi.org/10.1007/978-1-4899-7687-1_192}

\bibitem{adamopt}
Kingma, D.P., Ba, J.: Adam: A method for stochastic optimization.
\newblock arXiv preprint arXiv:1412.6980  (2014)

\bibitem{koller_deep_2018}
Koller, O., Zargaran, S., Ney, H., Bowden, R.: Deep {Sign}: {Enabling} {Robust}
  {Statistical} {Continuous} {Sign} {Language} {Recognition} via {Hybrid}
  {CNN}-{HMMs}.
\newblock International Journal of Computer Vision \textbf{126}(12), 1311--1325
  (2018).
\newblock \doi{10.1007/s11263-018-1121-3}.
\newblock \urlprefix\url{https://doi.org/10.1007/s11263-018-1121-3}

\bibitem{li_modout_2017}
Li, F., Neverova, N., Wolf, C., Taylor, G.: Modout: {Learning} {Multi}-{Modal}
  {Architectures} by {Stochastic} {Regularization}.
\newblock In: 2017 12th {IEEE} {International} {Conference} on {Automatic}
  {Face} {Gesture} {Recognition} ({FG} 2017), pp. 422--429 (2017).
\newblock \doi{10.1109/FG.2017.59}.
\newblock ISSN: null

\bibitem{liu_learning_2013}
Liu, L., Shao, L.: Learning discriminative representations from {RGB}-{D} video
  data.
\newblock In: Proceedings of the {Twenty}-{Third} international joint
  conference on {Artificial} {Intelligence}, {IJCAI} '13, pp. 1493--1500. AAAI
  Press, Beijing, China (2013)

\bibitem{cross}
Mannor, S., Peleg, D., Rubinstein, R.: The cross entropy method for
  classification.
\newblock In: Proceedings of the 22nd international conference on {Machine}
  learning, {ICML} '05, pp. 561--568. Association for Computing Machinery,
  Bonn, Germany (2005).
\newblock \doi{10.1145/1102351.1102422}.
\newblock \urlprefix\url{https://doi.org/10.1145/1102351.1102422}

\bibitem{paper_siu}
Mercanoglu~Sincan, O., Tur, A.O., Yalim~Keles, H.: Isolated {Sign} {Language}
  {Recognition} with {Multi}-scale {Features} using {LSTM}.
\newblock In: 2019 27th {Signal} {Processing} and {Communications}
  {Applications} {Conference} ({SIU}), pp. 1--4 (2019).
\newblock \doi{10.1109/SIU.2019.8806467}.
\newblock ISSN: 2165-0608

\bibitem{murakami_gesture_1991}
Murakami, K., Taguchi, H.: Gesture recognition using recurrent neural networks.
\newblock In: Proceedings of the {SIGCHI} conference on {Human} factors in
  computing systems {Reaching} through technology - {CHI} '91, pp. 237--242.
  ACM Press, New Orleans, Louisiana, United States (1991).
\newblock \doi{10.1145/108844.108900}.
\newblock \urlprefix\url{http://portal.acm.org/citation.cfm?doid=108844.108900}

\bibitem{neverova_moddrop_2016}
Neverova, N., Wolf, C., Taylor, G., Nebout, F.: {ModDrop}: {Adaptive}
  {Multi}-{Modal} {Gesture} {Recognition}.
\newblock IEEE Transactions on Pattern Analysis and Machine Intelligence
  \textbf{38}(8), 1692--1706 (2016).
\newblock \doi{10.1109/TPAMI.2015.2461544}.
\newblock Conference Name: IEEE Transactions on Pattern Analysis and Machine
  Intelligence

\bibitem{nunez_convolutional_2018}
Núñez, J.C., Cabido, R., Pantrigo, J.J., Montemayor, A.S., Vélez, J.F.:
  Convolutional {Neural} {Networks} and {Long} {Short}-{Term} {Memory} for
  skeleton-based human activity and hand gesture recognition.
\newblock Pattern Recognition \textbf{76}, 80--94 (2018).
\newblock \doi{10.1016/j.patcog.2017.10.033}.
\newblock
  \urlprefix\url{http://www.sciencedirect.com/science/article/pii/S0031320317304405}

\bibitem{nishida_multimodal_2016}
Nishida, N., Nakayama, H.: Multimodal {Gesture} {Recognition} {Using}
  {Multi}-stream {Recurrent} {Neural} {Network}.
\newblock In: T.~Bräunl, B.~McCane, M.~Rivera, X.~Yu (eds.) Image and {Video}
  {Technology}, Lecture {Notes} in {Computer} {Science}, pp. 682--694. Springer
  International Publishing, Cham (2016).
\newblock \doi{10.1007/978-3-319-29451-3_54}

\bibitem{pytorch}
Paszke, A., Gross, S., Chintala, S., Chanan, G., Yang, E., DeVito, Z., Lin, Z.,
  Desmaison, A., Antiga, L., Lerer, A.: Automatic differentiation in {PyTorch}.
\newblock In: NIPS-W (2017)

\bibitem{pigou2014sign}
Pigou, L., Dieleman, S., Kindermans, P.J., Schrauwen, B.: Sign language
  recognition using convolutional neural networks.
\newblock In: Workshop at the European Conference on Computer Vision, pp.
  572--578. Springer (2014)

\bibitem{pigou_beyond_2018}
Pigou, L., van~den Oord, A., Dieleman, S., Van~Herreweghe, M., Dambre, J.:
  Beyond {Temporal} {Pooling}: {Recurrence} and {Temporal} {Convolutions} for
  {Gesture} {Recognition} in {Video}.
\newblock International Journal of Computer Vision \textbf{126}(2-4), 430--439
  (2018).
\newblock \doi{10.1007/s11263-016-0957-7}.
\newblock \urlprefix\url{http://link.springer.com/10.1007/s11263-016-0957-7}

\bibitem{pisharady_recent_2015}
Pisharady, P.K., Saerbeck, M.: Recent methods and databases in vision-based
  hand gesture recognition: {A} review.
\newblock Computer Vision and Image Understanding \textbf{141}, 152--165
  (2015).
\newblock \doi{10.1016/j.cviu.2015.08.004}.
\newblock
  \urlprefix\url{http://www.sciencedirect.com/science/article/pii/S1077314215001794}

\bibitem{HMMIntro}
{Rabiner}, L., {Juang}, B.: An introduction to hidden markov models.
\newblock IEEE ASSP Magazine \textbf{3}(1), 4--16 (1986)

\bibitem{imagenet}
Russakovsky, O., Deng, J., Su, H., Krause, J., Satheesh, S., Ma, S., Huang, Z.,
  Karpathy, A., Khosla, A., Bernstein, M., et~al.: Imagenet large scale visual
  recognition challenge.
\newblock International Journal of Computer Vision \textbf{115}(3), 211--252
  (2015)

\bibitem{santos_dynamic_2020}
Santos, C.C.d., Samatelo, J.L.A., Vassallo, R.F.: Dynamic gesture recognition
  by using {CNNs} and star {RGB}: {A} temporal information condensation.
\newblock Neurocomputing  (2020).
\newblock \doi{10.1016/j.neucom.2020.03.038}.
\newblock
  \urlprefix\url{http://www.sciencedirect.com/science/article/pii/S092523122030391X}

\bibitem{pomegranate}
Schreiber, J.: Pomegranate: fast and flexible probabilistic modeling in python.
\newblock arXiv:1711.00137 [cs, stat]  (2018).
\newblock \urlprefix\url{http://arxiv.org/abs/1711.00137}.
\newblock ArXiv: 1711.00137

\bibitem{vgg}
Simonyan, K., Zisserman, A.: Very deep convolutional networks for large-scale
  image recognition.
\newblock arXiv preprint arXiv:1409.1556  (2014)

\bibitem{tsironi_gesture_2016}
Tsironi, E., Barros, P., Wermter, S.: Gesture {Recognition} with a
  {Convolutional} {Long} {Short}-{Term} {Memory} {Recurrent} {Neural}
  {Network}.
\newblock Computational Intelligence p.~6 (2016)

\bibitem{paper_eurocon}
Tur, A.O., Keles, H.Y.: Isolated {Sign} {Recognition} with a {Siamese} {Neural}
  {Network} of {RGB} and {Depth} {Streams}.
\newblock In: {IEEE} {EUROCON} 2019 -18th {International} {Conference} on
  {Smart} {Technologies}, pp. 1--6 (2019).
\newblock \doi{10.1109/EUROCON.2019.8861945}

\end{thebibliography}

\end{document}